\documentclass{article}
\usepackage[toc,page]{appendix}
\usepackage[square,sort,comma,numbers]{natbib}
\usepackage[final]{nips_2016}

\usepackage[utf8]{inputenc} 
\usepackage[T1]{fontenc}    
\usepackage{url}            
\usepackage{booktabs}       
\usepackage{amsfonts}       
\usepackage{nicefrac}       
\usepackage{microtype}      
\usepackage{graphicx}
\usepackage{color}
\usepackage{floatrow}
\usepackage{footnote}
\makesavenoteenv{tabular}
\usepackage[lofdepth,lotdepth,caption=false]{subfig}
\usepackage{amsmath}
\usepackage{titlesec}			
\title{Single-Image Depth Perception in the Wild}

\author{
	Weifeng Chen \qquad   Zhao Fu \qquad    Dawei Yang  \qquad   Jia Deng \\
	University of Michigan, Ann Arbor \\
	\texttt{\{wfchen,zhaofu,ydawei,jiadeng\}@umich.edu} \\
}

\newcommand{\smallpara}[1]{\noindent {\bf #1}}
\begin{document}

\maketitle

\begin{abstract}
This paper studies single-image depth perception in the wild, i.e., recovering depth from
a single image taken in
unconstrained settings. We introduce a new dataset ``Depth in the Wild'' consisting of images in the wild annotated with
relative depth between pairs of random points. We also propose a new algorithm that learns to
estimate metric depth using annotations of relative depth. Compared to the state of the
art, our algorithm is simpler and performs better. Experiments show that our algorithm,
combined with existing RGB-D data and our new relative depth annotations, significantly
improves single-image depth perception in the wild. 

\end{abstract}

\begin{figure}[h!]
	\centering
	\includegraphics[width=0.9\textwidth]{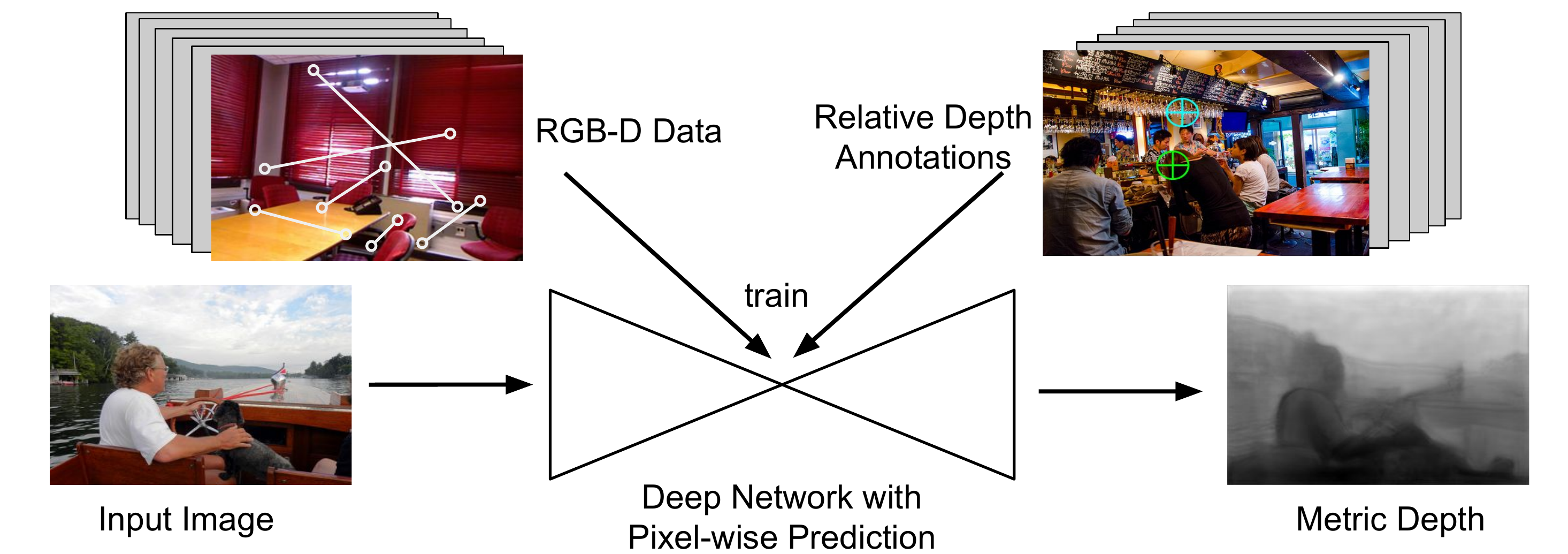}
	\caption{We crowdsource annotations of relative depth and train
          a deep network to recover depth from a single image taken in unconstrained
          settings (``in the wild''). }
	\label{fig:teaser}
\end{figure}

\section{Introduction}

Depth from a single RGB image is a fundamental problem in vision. Recent years
have seen rapid progress thanks to data-driven methods~\cite{Karsch:TPAMI:14,hoiem2005automatic,saxena2009make3d}, in
particular, deep neural
networks trained on large RGB-D
datasets~\cite{silberman2012indoor,geiger2013vision,liu2015deep,ladicky2014pulling,eigen2015predicting,baig2015coupled,li2015depth}. But
such advances have yet to broadly impact higher-level tasks. One reason is that many higher-level
tasks must operate on images ``in the wild''---images taken with no constraints on cameras, locations,
scenes, and objects---but the RGB-D datasets used to train and evaluate image-to-depth
systems are constrained in one way or another.

\begin{figure}[t]
	\centering
	\includegraphics[width=0.9\textwidth]{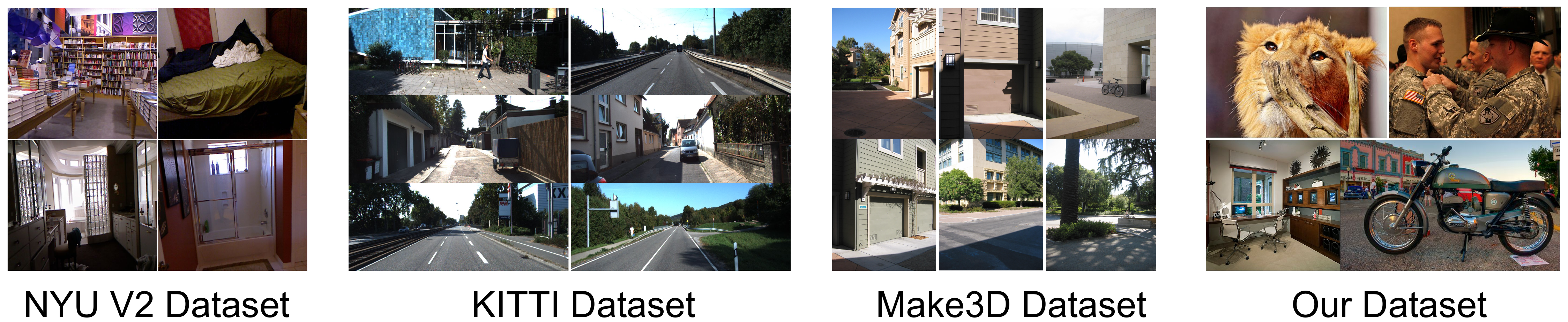}
	\caption{Example images from current RGB-D datasets and our Depth in the Wild
          (DIW) dataset.}
	\label{fig:dataset_comparison}
\end{figure}

Current RGB-D datasets were collected by depth
sensors~\cite{silberman2012indoor,geiger2013vision}, which are limited in range and
resolution, and often fail on specular or transparent objects~\cite{chiu2011improving}. In
addition, because there is no Flickr for RGB-D
images, researchers have to manually capture the images. As a
result, current RGB-D datasets are limited in the diversity of scenes. For example, NYU
depth~\cite{silberman2012indoor} consists mostly of indoor scenes with no human
presence; KITTI~\cite{geiger2013vision} consists mostly of road scenes captured from a
car; Make3D~\cite{saxena2009make3d,saxena2005learning} consists mostly of outdoor scenes of
the Stanford campus (Figure.~\ref{fig:dataset_comparison}). While these datasets are pivotal in driving research, it is unclear whether systems trained on them can generalize
to images in the wild.

Is it possible to collect ground-truth depth for images in the wild? Using depth sensors
in unconstrained settings is not yet feasible. Crowdsourcing seems viable, but humans are not good at estimating
metric
depth, or 3D metric structure in general~\cite{todd2003visual}.  In fact, metric depth from a single image is fundamentally
ambiguous: a tree behind a house can be slightly
bigger but further away, or slightly smaller but closer---the absolute depth difference
between the house and the tree cannot be uniquely determined. 
Furthermore, even in cases where humans can estimate metric
 depth, it is unclear how to elicit the values from them. 

But humans are better at judging relative depth~\cite{todd2003visual}: ``Is point A closer than
point B?'' is often a much easier question for humans. Recent work by Zoran et
al.~\cite{zoran2015learning} shows that it is possible to learn to estimate metric depth using only
annotations of relative depth. Although such metric depth estimates are only accurate up to
monotonic transformations, they may well be sufficiently useful for high-level tasks,
especially for occlusion reasoning.  
The seminal results by Zoran et al.\@ point to two fronts for further progress: (1) collecting a large amount of relative depth
annotations for images in the wild and (2) improving the algorithms that learn from
annotations of relative depth. 

In this paper, we make contributions on both fronts. Our first contribution is a new dataset called
``Depth in the Wild'' (DIW). It consists of 495K diverse images, each annotated with
randomly sampled points and their relative depth. We sample one pair of points per
image to minimize the redundancy of annotation~\footnote{A small percentage of
  images have duplicates and thus have multiple pairs. }. To the best of our knowledge this is the first large-scale dataset consisting of images in the wild with relative depth annotations. We
demonstrate that this dataset can be used as an evaluation benchmark 
as well as a training resource~\footnote{Project website: \url{http://www-personal.umich.edu/~wfchen/depth-in-the-wild}.}. 

Our second contribution is a new algorithm for learning to estimate metric depth using only
annotations of relative depth. Our algorithm not only significantly outperforms that of Zoran et
al.~\cite{zoran2015learning}, but is also simpler. The algorithm of Zoran et al.~\cite{zoran2015learning} first 
learns a classifier to predict the ordinal relation between two points in an
image. Given a new image, this classifier is repeatedly applied to predict the ordinal
relations between a sparse set of point pairs (mostly between the
centers of neighboring superpixels). The algorithm then reconstructs depth from the predicted ordinal
relations by solving a constrained quadratic optimization that enforces
additional smoothness 
constraints and reconciles
potentially inconsistent ordinal relations. Finally, the algorithm estimates depth for all pixels assuming a constant
depth within each superpixel. 

In contrast, our algorithm consists of a single deep network that directly predicts
pixel-wise depth (Fig.~\ref{fig:teaser}). The network takes an entire image as input, consists of off-the-shelf components, and can be trained entirely with annotations of
relative depth. The novelty of our approach lies in the combination of two
ingredients:
(1) a multi-scale deep network that produces pixel-wise prediction of metric depth and (2) a
loss function using relative depth. Experiments show that our method produces pixel-wise depth
that is more accurately ordered, outperforming not only the method by Zoran et
al.~\cite{zoran2015learning} but also the state-of-the-art image-to-depth system by Eigen
et al.~\cite{eigen2015predicting} trained with ground-truth metric depth. Furthermore,
combing our new algorithm, our new dataset, and existing RGB-D data significantly improves
single-image depth estimation in the wild. 

\vspace{-2 mm}
\section{Related work}
\vspace{-2 mm}

\smallpara{RGB-D Datasets:} 
Prior work on constructing RGB-D datasets has relied on either
Kinect~\cite{janoch2013category,silberman2012indoor,song2015sun,choi2016large} or
LIDAR~\cite{geiger2013vision,saxena2009make3d}. Existing Kinect-based datasets are limited to
indoor scenes; existing LIDAR-based datasets are biased towards scenes of man-made
structures~\cite{geiger2013vision,saxena2009make3d}.
In contrast, our dataset covers a much wider variety of
scenes; it can be easily expanded with large-scale crowdsourcing and the virually
umlimited Internet images. 

\smallpara{Intrinsic Images in the Wild:} Our work draws inspiration from Intrinsic Images in the Wild~\cite{bell2014intrinsic},
a seminal work that crowdsources annotations of relative reflectance on unconstrained
images. Our work differs in goals as well as in several design decisions. 
First, we sample random points instead of centers of superpixels, because unlike
reflectance, it is unreasonable to assume a 
constant depth within a superpixel. Second, we sample only one pair of points per
image instead of many to maximize the value of human annotations. 

\smallpara{Depth from a Single Image:}
Image-to-depth is a long-standing problem with a large body of
literature~\cite{BarronTPAMI2015,saxena20083,saxena2005learning,Karsch:TPAMI:14,liu2015deep,ladicky2014pulling,eigen2015predicting,baig2015coupled,li2015depth,BarronTPAMI2015,xiong2015shading,hane2015direction,liu2010single,shelhamer2015scene,shi2015break,zhuo2015indoor}. The recent convergence of deep neural networks and RGB-D
datasets~\cite{silberman2012indoor,geiger2013vision} has led to major advances~\cite{zhang2015monocular,liu2015deep,wang2015towards,eigen2015predicting,li2015depth,zoran2015learning}. But the networks in
these previous works, with the exception of ~\cite{zoran2015learning}, were trained
exclusively using ground-truth metric depth, whereas our approach uses relative depth. 

Our work is inspired by that of Zoran et al.~\cite{zoran2015learning}, which proposes to use
a deep network to repeatedly classify pairs of points sampled based on superpixel
segmentation, and to reconstruct per-pixel metric depth by solving an additional optimization
problem. Our approach is different: it consists of a single deep
network trained end-to-end that directly predicts per-pixel metric depth; there is no
intermediate classification of ordinal relations and as a result no
optimization needed to resolve inconsistencies. 

\smallpara{Learning with Ordinal Relations:} 
Several recent works~\cite{zhou2015learning,narihira2015learning} have used the ordinal
relations from the Intrinsic Images in
the Wild dataset~\cite{bell2014intrinsic} to estimate surface refletance. Similar to Zoran et
al.~\cite{zoran2015learning},  Zhou et al.~\cite{zhou2015learning} first learn a
deep network to classify the ordinal relations between pairs of points and then make them
globally consistent through energy minimization. 

Narihira et al.~\cite{narihira2015learning} learn a ``lightness potential'' network that takes an image patch and predicts the metric reflectance of the center pixel. But this network is applied to only a sparse set of
 pixels. Although in principle this lightness potential network can be
applied to every pixel to produce pixel-wise reflectance, doing so would be quite expensive.
Making it fully convolutional (as the authors mentioned in~\cite{narihira2015learning})
only solves it partially: as long as the lightness potential network has downsampling layers, which is the case
in~\cite{narihira2015learning}, the final output will be downsampled
accordingly. Additional resolution augmentation (such as the ``shift and stitch'' approach
~\cite{sermanet2013overfeat}) is thus needed. In contrast, our approach
completely avoids such issues and directly outputs pixel-wise estimates.

Beyond intrinsic images, ordinal relations have been used widely in computer vision and
machine learning, including object recognition~\cite{parikh2011relative} and learning to rank~\cite{cao2007learning,joachims2002optimizing}.

\section{Dataset construction}

 \begin{figure}[t]
 	\RawFloats
 	\centering
 	\begin{minipage}[t]{0.43\textwidth}
 		\centering
 
 		\includegraphics[width=0.83\textwidth]{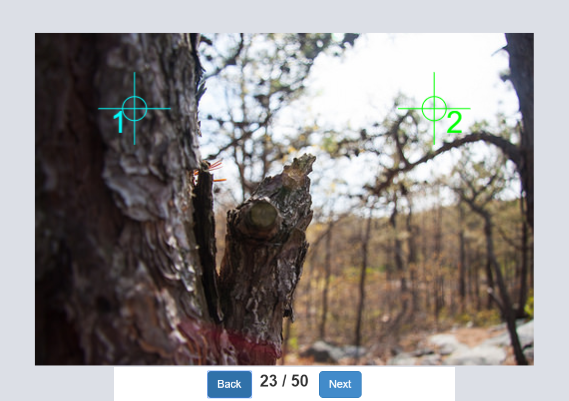}
 		\caption{ Annotation UI. The user presses '1' or '2' to pick the closer point. }
 		\label{fig:gui}
 	\end{minipage}\quad
 	\begin{minipage}[t]{0.53\textwidth}
 		\centering
 		\includegraphics[width=0.674\textwidth]{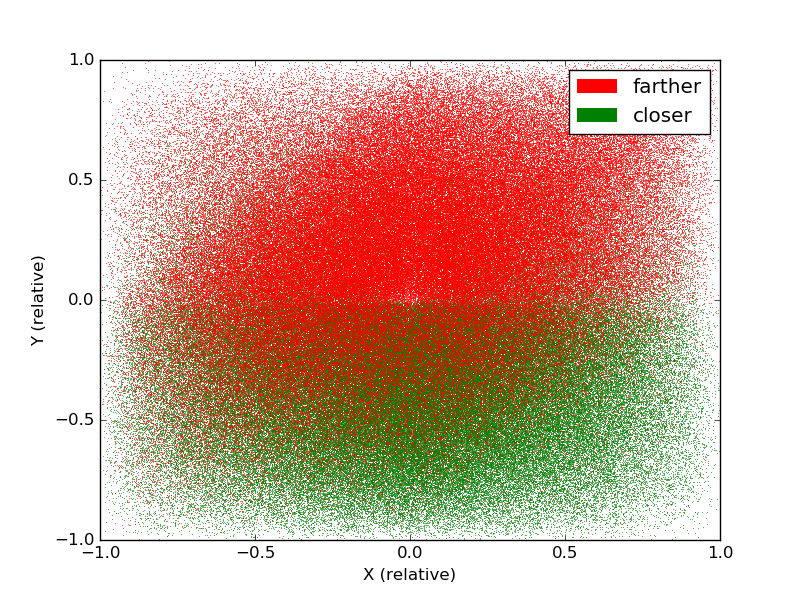}
 
 		\caption{ Relative image location (normalized to [-1,1]) and relative depth of two random points. }
 		\label{fig:bias_graph}
 	\end{minipage}
 \end{figure}
We gather images from Flickr. We use random query keywords sampled from an English
dictionary and exclude artificial images such as drawings and clip arts. 
To collect annotations of relative depth, we present a crowd worker an image and two
highlighted points (Fig.~\ref{fig:gui}), and ask ``which point is closer, point 1, point
2, or hard to tell?'' The worker presses a key to respond.

\smallpara{How Many Pairs?} How many pairs of points should we query per image? We sample just one per image
because this maximizes the amount of information from human annotators. Consider
the other extreme---querying all possible pairs of points in the same image. This is
wasteful because pairs of points in close proximity are likely to have the same relative
depth. In other words, querying one more pair from the same image may add less information
than querying one more pair from a new image. Thus querying only one pair per image is
more cost-effective. 

\smallpara{Which Pairs?} Which two points should we query given an image? The simplest way would be to 
sample two random points from the 2D plane. But this results in a severe bias that can
be easily exploited: if an algorithm simply classifies the lower point in the image to be closer in depth, it will agree with humans $85.8\%$ of the
time (Fig.~\ref{fig:bias_graph}). Although this bias is natural, it makes the dataset less useful
as a benchmark.

\begin{figure}[t]
	\centering
	\includegraphics[width=1.0\textwidth]{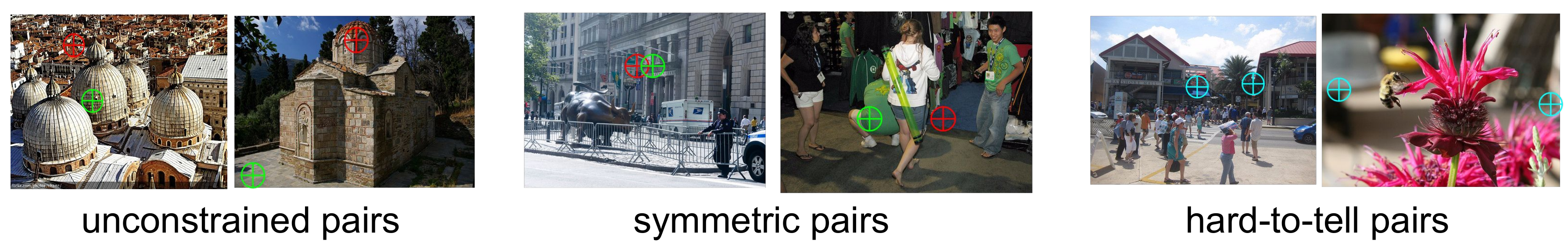}
	\caption{Example images and annotations. Green points are those annotated as closer in depth. }
	\label{fig:points_in_our_dataset}
\end{figure}

An alternative is to sample two points uniformly from
a random horizontal line, which makes it impossible
to use the $y$ image coordinate as a cue. But we find yet another bias: if an algorithm simply
classifies the point closer to the center of the image to be closer in depth, it will
agree with humans $71.4\%$ of the time. This leads to a third approach: uniformly
sample two \emph{symmetric} points with respect to the center from a random horizontal
line (the middle column of Fig.~\ref{fig:points_in_our_dataset}). With the symmetry
enforced, we are not able to find a simple yet effective rule based purely on
image coordinates: the left point is almost equally likely ($50.03\%$) to be closer than
the right one. 

Our final dataset consists of a roughly 50-50 combination of unconstrained pairs and symmetric
pairs, which strikes a balance between the need for representing natural scene statistics and the 
need for performance differentiation.

\smallpara{Protocol and Results:}
We crowdsource the annotations using Amazon Mechanical Turk (AMT). To remove spammers, we
insert into all tasks gold-standard images verified by ourselves, and reject workers whose
accumulative accuracy on the gold-standard images is below 85\%. 
We assign each query (an image and a point pair) to two workers, and add the query to our
dataset if both workers can tell the relative depth and agree with each other; otherwise
the query is discarded. Under this
protocol, the chance of adding a wrong answer to our dataset is less than 1\% as measured on
the gold-standard images. 

We processed 1.24M images on AMT and obtained 0.5M valid answers (both
workers can tell the relative depth and agree with each other). Among the valid answers,
261K are for unconstrained pairs and 240K are for symmetric pairs. For unconstrained
pairs, It takes a median of 3.4
seconds for a worker to decide, and two workers agree on the relative depth 
52\% of the time; for symmetric pairs, the numbers are 3.8s and 32\%. These
numbers suggest that the symmetric pairs are indeed harder. Fig.~\ref{fig:points_in_our_dataset} presents 
examples of different kinds of queries.

\section{Learning with relative depth}

How do we learn to predict metric depth given only annotations of relative depth? Zoran et al.~\cite{zoran2015learning}
first learn a classifier to predict ordinal relations between centers of superpixels, and then
reconcile the relations to recover depth using energy minimization, and then interpolate
within each superpixel to produce per-pixel depth.

We take a simpler approach. The idea is that any image-to-depth algorithm
would have to compute a function that maps an image to pixel-wise depth. Why not
represent this function as a neural network and learn it from end to end? We just need two ingredients: (1)
a network design that outputs the same resolution as the input, and (2) a way
to train the network with annotations of relative depth. 

\smallpara{Network Design:}
Networks that output the same resolution as the input are aplenty, including the recent designs for depth
estimation~\cite{eigen2015predicting,eigen2014depth} and those for semantic
segmentation~\cite{long2015fully} and edge detection~\cite{DBLP:journals/corr/XieT15}. A
common element is processing and passing
information across multiple scales. 

\begin{figure}[t]
	\centering
	\includegraphics[width=0.9\textwidth]{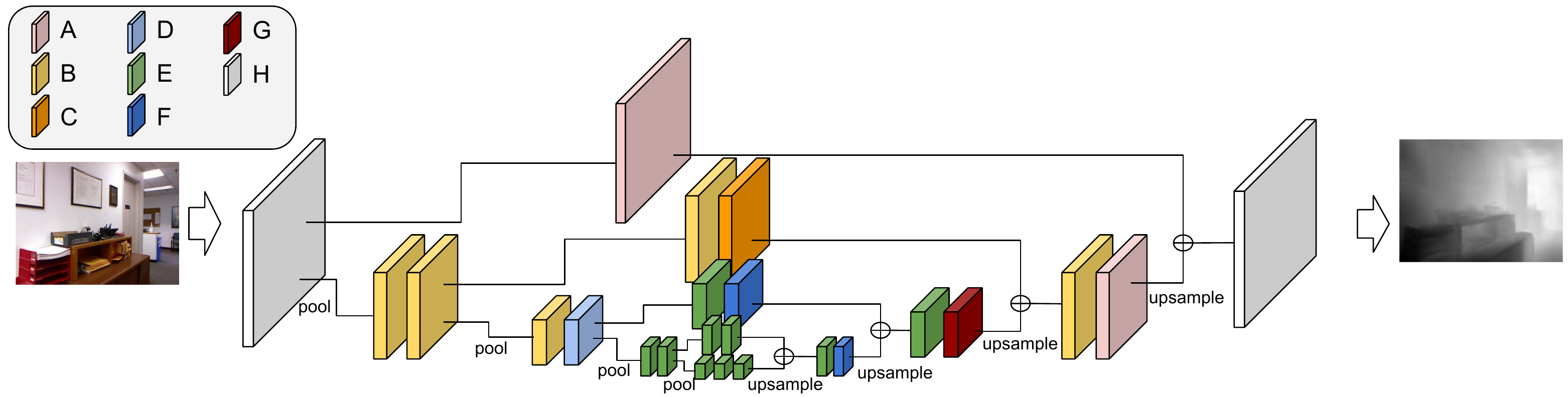}
	\caption{ Network design. Each block represents a layer. Blocks sharing the
          same color are identical. The $\oplus$ sign denotes the element-wise addition. Block H is a convolution
          with 3x3 filter. All other blocks denote the Inception module shown in
          Figure~\ref{fig:inception}. Their parameters are detailed in
          Tab.~\ref{table:network_parameters}}

	\label{fig:network}
\end{figure}

In this work, we use a variant of the recently introduced ``hourglass''
network (Fig.~\ref{fig:network}), which has been used to achieve state-of-the-art results on human pose
estimation~\cite{newell2016stacked}. It consists of a series of
convolutions (using a variant of the inception~\cite{szegedy2015going} module)
and downsampling, followed by a series of convolutions and upsampling, interleaved with
skip connections that add back features from high resolutions. The symmetric shape of
the network resembles a ``hourglass'', hence the name. We refer the reader
to~\cite{newell2016stacked} for comparing the design to related work. For our purpose, 
this particular choice is not essential, as the various designs mainly differ in
how information from different scales is dispersed and aggregated, and it is 
possible that all of them can work equally well for our task. 

\smallpara{Loss Function:} How do we train the network 
using only ordinal annotations? All we need is a loss function 
that encourages the predicted depth map to agree with the ground-truth ordinal
relations. Specifically, consider a training image $I$ and its $K$ queries $R = \{(i_k,j_k,r_k)\},
k=1,\ldots,K$, where $i_k$ is the location of the first point in the $k$-th query, $j_k$ is
the location of the second point in the $k$-th query, and $r_k \in \{+1,-1, 0\}$ is the ground-truth depth relation between $i_k$ and
$j_k$: closer ($+1$), further ($-1$), and equal ($0$). Let $z$ be the predicted depth map
and $z_{i_k}, z_{j_k}$ be the depths at point $i_k$ and  $j_k$. We define a loss function
\begin{equation}
L(I, R, z) = \sum_{k=1}^K \psi_k(I,i_k,j_k,r, z), 
\end{equation}
where  $\psi_k(I,i_k,j_k, z)$ is the loss for the $k$-th query
\begin{align}
\psi_k(I,i_k,j_k, z) =\left\{
\begin{array}{ll}
\log\left(1+\exp(-z_{i_k} + z_{j_k}) \right), & r_k = +1 \\
\log\left(1+\exp(z_{i_k} - z_{j_k}) \right),  &  r_k = -1 \\
 (z_{i_k} - z_{j_k})^2, &  r_k = 0. \\
\end{array}
\right.
\label{equation:definition_of_loss1}
\end{align}
This is essentially a ranking loss: it encourages a small difference between depths if the
ground-truth relation is equality; otherwise it encourages a large difference.

\smallpara{Novelty of Our Approach:} Our novelty lies in the combination
of a deep network that does pixel-wise prediction and a ranking loss placed on the pixel-wise
prediction. A deep network that does pixel-wise prediction is not new, nor is a ranking loss. But
to the best of our knowledge, such a combination has not been proposed before, and in
particular not for estimating depth.

\begin{figure}
\CenterFloatBoxes
\begin{floatrow}
\ffigbox[\FBwidth]
  {\includegraphics[width=0.3\textwidth]{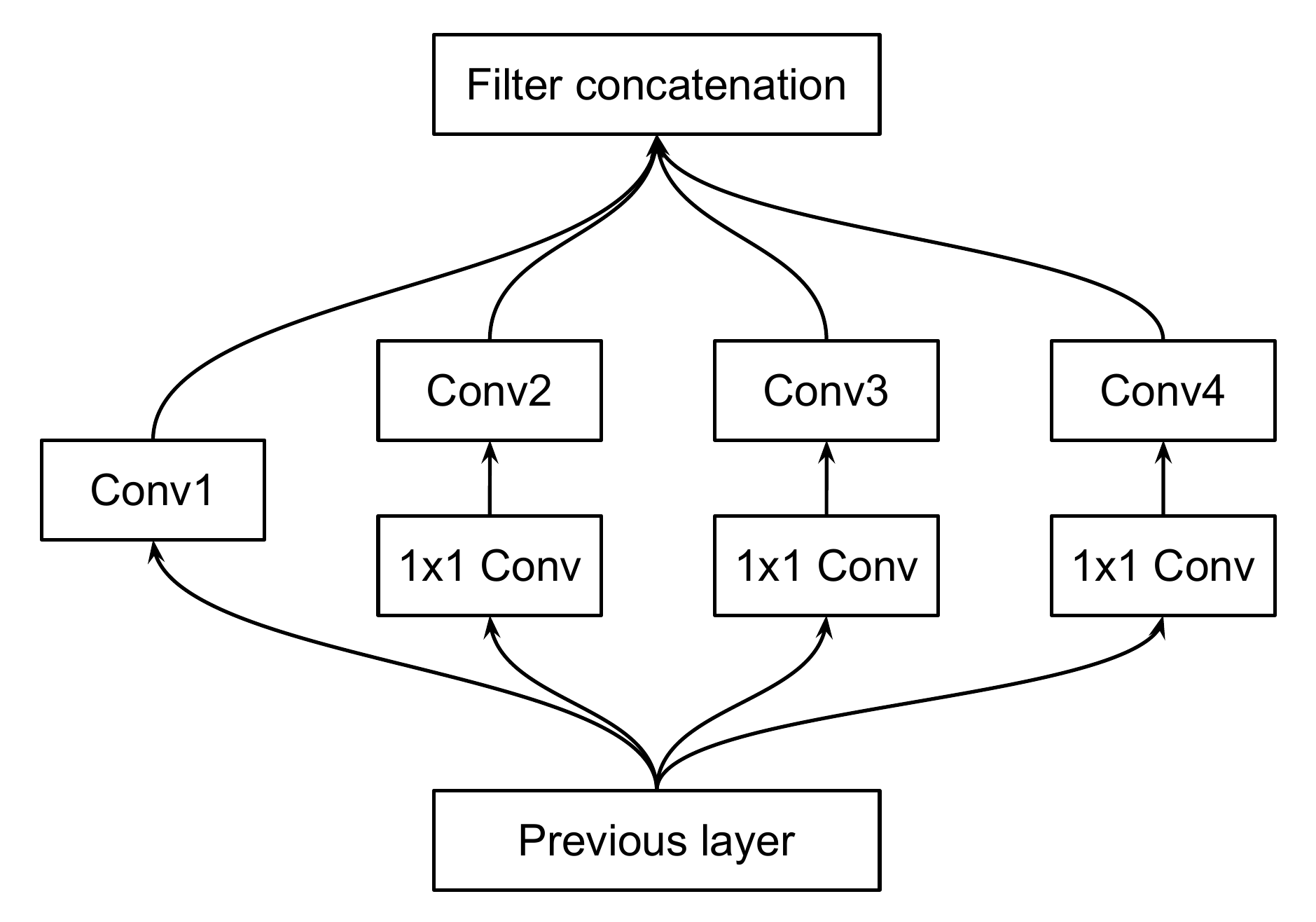}}
  {\caption{Variant of Inception Module~\cite{szegedy2015going} used by us.}
  \label{fig:inception}}
\killfloatstyle
\ttabbox
  {
  \resizebox{0.6\textwidth}{!}{
\begin{tabular}{ c  c  c  c  c  c  c  c  } 
					\toprule
					Block Id &	A     &   B      &C          &D          &E          &F          &G          		\\
					\midrule
					\#In/\#Out& 128/64&	128/128&	128/128&	128/256&	256/256&	256/256&	256/128\\
					
					Inter Dim& 64 &    32 &    64 &    32 &     32&     64 & 32 \\
					
					Conv1&   1x1&	1x1&	1x1&	1x1&	1x1&	1x1&	1x1 	\\
					
					Conv2&   3x3&	3x3&	3x3&	3x3&	3x3&	3x3	&3x3	\\
					
					Conv3&   7x7&	5x5&	7x7&	5x5&	5x5&	7x7&	5x5	\\
					
					Conv4& 11x11&	7x7&	11x11&	7x7&	7x7&	11x11&	7x7	\\
					
					\bottomrule
				\end{tabular}}
  }
  {\caption{Parameters for each type of layer in our network. \textit{Conv1} to \textit{Conv4} are sizes of the filters used in the components of Inception module shown in Figure.\ref{fig:inception}. Conv2 to 4 share the same number of input and is specified in \textit{Inter Dim}.}
  \label{table:network_parameters}}
\end{floatrow}
\end{figure}

\begin{figure}[t] 
	\centering
	\includegraphics[width=1.0\textwidth]{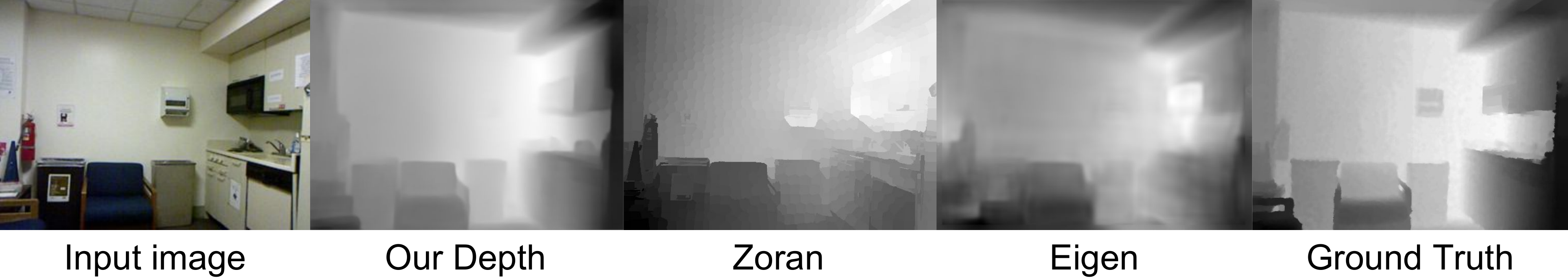}
	\caption{Qualitative results on NYU Depth by our method, the method of Eigen et
          al.~\cite{eigen2015predicting}, and the method of Zoran et
          al.~\cite{zoran2015learning}. All depth maps except ours are directly from
          \cite{zoran2015learning}. More results are in the Appendix.}
	\label{fig:qual_nyu}
\end{figure}

\section{Experiments on NYU Depth}
We evaluate our method using NYU Depth~\cite{silberman2012indoor}, which consists of indoor scenes with ground-truth Kinect depth. We use the same setup as that of Zoran et
al.~\cite{zoran2015learning}: point pairs are sampled from the training images (the subset
of NYU Depth consisting of 795 images with semantic labels) using
superpixel segmentation and their ground-truth ordinal relations are generated by comparing
the ground-truth Kinect depth; the same procedure is applied to the test set to generate
the point pairs for evaluation (around 3K pairs per image). 
We use the same training and test data as Zoran et al.~\cite{zoran2015learning}. 

\begin{savenotes}
\begin{table}[!htbp]
	\centering
	\begin{center}
		\resizebox{0.40\textwidth}{!}{\begin{tabular}{ l  c  c  c } 
			\toprule
			Method & WKDR & $\text{WKDR}^{=}$ & $\text{WKDR}^{\neq}$ \\ 
			\midrule
			Ours &  {\bf 35.6\% }& {\bf 36.1\% } & {\bf 36.5\%} \\

			Zoran~\cite{zoran2015learning}	&  43.5\% &	44.2\% & 41.4\% \\

\midrule
			rand\_12K  &  {\bf 34.9\% } & {\bf 32.4\%} & {\bf 37.6\%}\\
			rand\_6K  &   36.1\% & 32.2\% & 39.9\%\\
			rand\_3K  &   35.8\% & 28.7\% & 41.3\%\\

\midrule

			Ours\_Full & \textbf{28.3\%} & \textbf{30.6\%} & \textbf{28.6\%} \\ 
			Eigen(A)~\cite{eigen2015predicting} &  37.5\% & 46.9\% &
			32.7\% \\
			Eigen(V)~\cite{eigen2015predicting} &  34.0\%& 43.3\% &  29.6\% \\
			\bottomrule
		\end{tabular}}
		\resizebox{0.4\textwidth}{!}{\begin{tabular}{  l  c  c  c  c  c  } 
			\toprule
			Method & RMSE & RMSE & RMSE~\footnote{Computed using our own implementation based on the definition given
			in~\cite{eigen2014depth}.} & absrel & sqrrel \\
			       &      & (log)&(s.inv) &    &         \\
 			\midrule
 			Ours & 1.13  &  0.39 & 0.26  & 0.36  &  0.46 \\
			Ours\_Full & 1.10 & 0.38 & 0.24 & 0.34 & 0.42 \\
			Zoran~\cite{zoran2015learning} & 1.20 & 0.42 & - & 0.40  & 0.54 \\
			Eigen(A)~\cite{eigen2015predicting} & 0.75 & 0.26 & 0.20 & 0.21 & 0.19 \\
			Eigen(V)~\cite{eigen2015predicting} & \textbf{0.64} & \textbf{0.21} & \textbf{0.17} & \textbf{0.16} & \textbf{0.12} \\
			Wang~\cite{wang2015towards} & 0.75 & - & - & 0.22 & - \\
			Liu~\cite{liu2015deep} & 0.82 & - & - & 0.23& -\\
			Li~\cite{li2015depth} & 0.82 & - & - & 0.23 & -\\
			Karsch~\cite{Karsch:TPAMI:14} & 1.20 & - & - & 0.35 & - \\
			Baig~\cite{baig2014im2depth} & 1.0 & - & - & 0.3 & -\\
			\bottomrule
		\end{tabular}}
	\end{center}
	\caption{Left table: ordinal error measures (disagreement rate with ground-truth depth ordering) on
          NYU Depth. Right able: metric error measures on NYU Depth. Details for each metric can
          be found in ~\cite{eigen2015predicting}. There are two versions of results by
          Eigen et al.~\cite{eigen2015predicting}, one using AlexNet (Eigen(A)) and one
          using VGGNet (Eigen(V)). Lower is better for all error
          measures. }

	\label{table:nyu_performance}
\end{table}
\end{savenotes}

As the system by Zoran et al.~\cite{zoran2015learning}, our network predicts one of the three ordinal
relations on the test pairs: equal ($=$), closer ($<$), or farther ($>$). We report WKDR, the weighted disagreement rate between the predicted ordinal relations and
ground-truth ordinal relations~\footnote{WKDR stands for ``Weighted Kinect Disagreement
  Rate''; the weight is set to 1 as in~\cite{zoran2015learning}}. We also report
$\text{WKDR}^{=}$ (disagreement rate on pairs whose ground-truth relations are $=$) and
$\text{WKDR}^{\neq}$ (disagreement rate on pairs whose ground-truth relations are $<$ or $>$). 

Since two ground-truth depths are almost never exactly the same, there needs to be a
relaxed definition of equality. Zoran et al.~\cite{zoran2015learning} define two points to have equal depths
if the ratio between their ground-truth depths is
within a pre-determined range. Our network predicts an equality relation if the depth difference is smaller
than a threshold $\tau$. The choice of this threshold will result in different values for
the error metrics ($\text{WKDR}$, $\text{WKDR}^{=}$, $\text{WKDR}^{\neq}$): if $\tau$ is
too small, most pairs will be predicted to be unequal and the error metric on equality relations ($\text{WKDR}^{=}$) will
be large; if $\tau$ is too big, most pairs will be predicted to be equal and the error
metric on inequality relations ($\text{WKDR}^{\neq}$) will be large. We choose the threshold $\tau$ that minimizes the maximum of the three
error metrics on a validation set held out from the training set. Tab.~\ref{table:nyu_performance} compares our network (\emph{ours}) versus that of
Zoran et al.~\cite{zoran2015learning}. Our network is trained with the same
data~\footnote{The code released by Zoran et al.~\cite{zoran2015learning} indicates that they train with a random subset of 800
  pairs per image instead of all the pairs. We follow the same procedure and only
  use a random subset of 800 pairs per image.} but
outperforms ~\cite{zoran2015learning} on all three metrics. 

Following~\cite{zoran2015learning}, we also compare with the state-of-art image-to-depth
system by Eigen et al.~\cite{eigen2015predicting}, which is trained on pixel-wise ground-truth metric
depth from the full NYU Depth training set (220K images). To compare fairly, we give our
network access to the full NYU Depth training set. In addition, we remove the limit of 800
point pairs per training image placed by Zoran et al and use all available pairs. The results
in Tab.~\ref{table:nyu_performance} show that our network (\emph{ours\_full}) achieves superior performance 
in estimating depth ordering. Granted, this comparison is
not entirely fair because ~\cite{eigen2015predicting} is not optimized for predicting
ordinal relations. But this comparison is still significant in that it shows that we can 
train on only relative depth and rival the state-of-the-art system in estimating depth up
to monotonic transformations. 

In Figure.~\ref{fig:qual_nyu} we show qualitative results on the same example images used
by Zoran et al.~\cite{zoran2015learning}. We see that although imperfect, the recovered
metric depth by our method is overall reasonable and qualitatively similar to that by the 
state-of-art system~\cite{eigen2015predicting} trained on ground-truth metric depth.

\smallpara{Metric Error Measures.} 
Our network is trained with relative depth, so it is unsurprising that it does
well in estimating depth up to ordering. But how good is the estimated depth in terms of
metric error? We thus evaluate conventional error
measures such as RMSE (the root mean squared error), which compares the absolute depth values to the
ground truths. Because our network is trained only on relative
depth and does not know the range of the ground-truth depth values, to make these error measures meaningful we
normalize the depth predicted by our network such that the mean and standard deviation are
the same as those of the mean depth map of the training
set. Tab.~\ref{table:nyu_performance} reports the results. We see that under these metric
error measures our network still outperforms the method of Zoran et
al.~\cite{zoran2015learning}. In addition, while our metric error is worse than the current
state-of-the-art, it is comparable to some of the earlier
methods (e.g.~\cite{Karsch:TPAMI:14}) that have access to 
ground-truth metric depth.

\begin{figure}[t]
	\centering
	\includegraphics[width=0.9\textwidth]{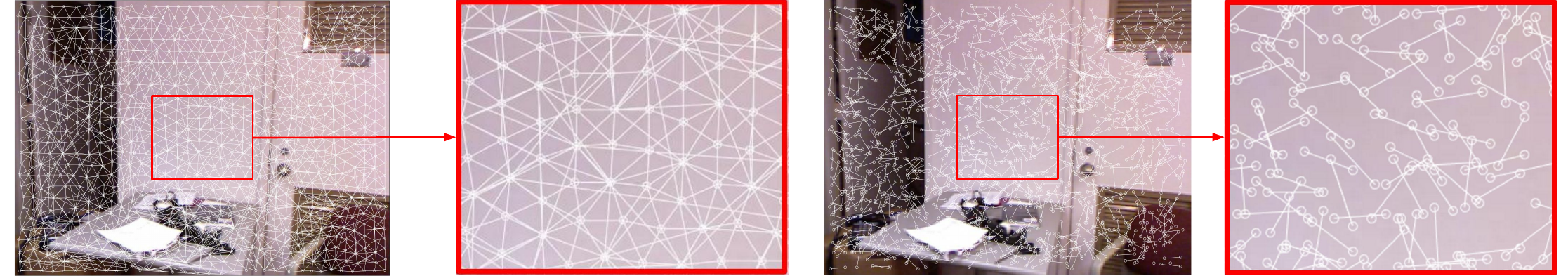}
	\caption{Point pairs generated through superpixel
          segmentation~\cite{zoran2015learning} (left) versus point pairs generated
          through random sampling with distance constraints (right). }
	\label{fig:random_structured_point}
\end{figure}

\smallpara{Superpixel Sampling versus Random Sampling.}
To compare with the method by Zoran et al.~\cite{zoran2015learning}, we train our network
using the same point pairs, which are pairs of centers of superpixels
(Fig.~\ref{fig:random_structured_point}). But is superpixel segmentation necessary? That
is, can we
simply train with randomly sampled points? 

To answer this question, we train our network with randomly sampled points. We 
constrain the distance between the two points to be between 13 and 19 pixels
(out of a 320$\times$240 image) such that the distance is similar to that between the
centers of neighboring superpixels. The results are included in Tab.~\ref{table:nyu_performance}. We see that
using 3.3k pairs per image (\textit{\textbf{rand\_3K}}) already
achieves comparable performance to the method by Zoran et
al.~\cite{zoran2015learning}. Using twice or four times as many pairs
(\textit{\textbf{rand\_6K}}, \textit{\textbf{rand\_12K}}) further improves
performance and significantly outperforms ~\cite{zoran2015learning}. 


It is worth noting that in all these experiments the test pairs are still from 
superpixels, so training on random pairs incurs a mismatch between
training and testing distributions. Yet we can still achieve comparable performance
despite this mismatch. This shows that our method can indeed operate without superpixel
segmentation.

\section{Experiments on Depth in the Wild}
In this section we experiment on our new Depth in the Wild (DIW) dataset. We split the
dataset into 421K training images and 74K test images~\footnote{4.38\% of images are duplicates downloaded using different query
  keywords and have more than one pairs of points. We have removed test images that have duplicates in the training set.}.

We report the WHDR (Weighted Human Disagreement Rate)~\footnote{All weights are 1. A
  pair of points can only have two possible ordinal relations (farther or closer) for DIW.} of 5 methods in
Tab.~\ref{table:amt_performance}: (1) the state-of-the-art system by Eigen et
al.~\cite{eigen2015predicting} trained on full NYU Depth; (2) our network
trained on full NYU Depth (Ours\_Full); (3) our network pre-trained on 
full NYU Depth and fine-tuned on DIW (Ours\_NYU\_DIW); (4) our network trained from
scratch on DIW (Ours\_DIW); (5) a baseline method that uses only the location of
the query points: classify the lower point to be closer or guess randomly if the two
points are at the same height (Query\_Location\_Only). 

\begin{figure}[t]
	\centering
	\includegraphics[width=1.0\textwidth]{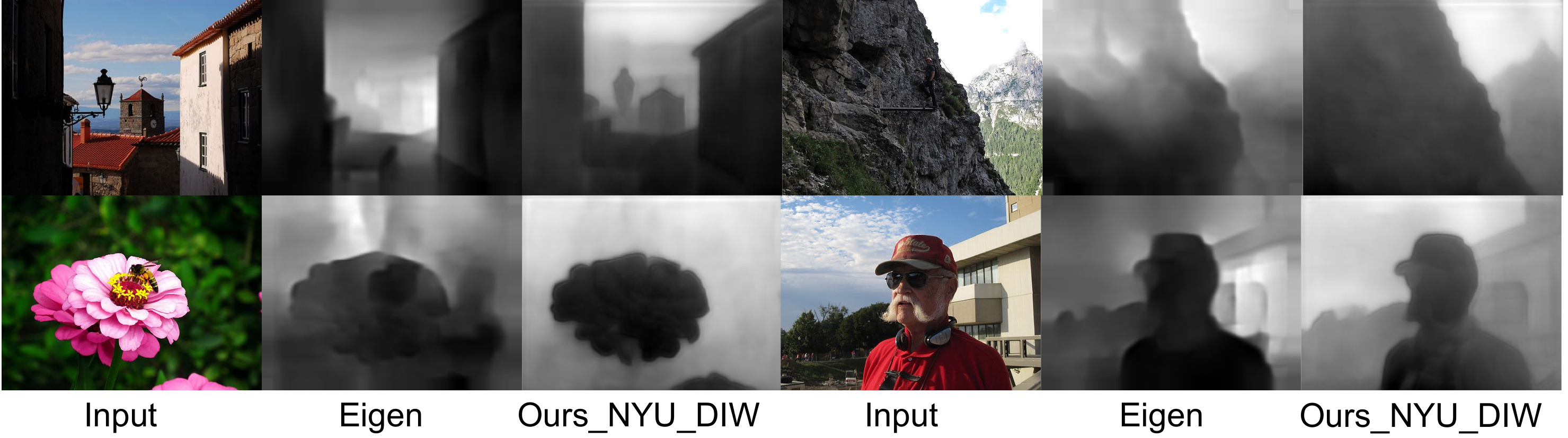}
	\caption{Qualitative results on our Depth in the Wild (DIW) dataset by our method and the method of Eigen et al.~\cite{eigen2015predicting}. More results are in the Appendix.}
	\label{fig:qual_DIW}
\end{figure}

We see that the best result is achieved by pre-training on NYU Depth and fine-tuning 
on DIW. Training  only on NYU Depth (Ours\_NYU and Eigen) does not work as well, which is 
expected because NYU Depth only has indoor scenes. 
Training from scratch on DIW
 achieves slightly better performance than those trained on only NYU Depth despite using much less 
 supervision. Pre-training on NYU Depth and fine-tuning on DIW leaverages all available
 data and achieves the best performance. As shown in Fig.~\ref{fig:qual_DIW}, the quality of predicted depth is
notably better with fine-tuning on DIW, especially for outdoor scenes. These results suggest that it is promising to combine existing  
 RGB-D data and crowdsourced annotations to advance the state-of-the art 
 in single-image depth estimation. 

\begin{table}[t]
	\centering
	\begin{center}
		\begin{tabular}{  c  c  c  c  c  c } 
			\hline
			Method & Eigen(V)~\cite{eigen2015predicting}& Ours\_Full & Ours\_NYU\_DIW & Ours\_DIW & Query\_Location\_Only \\
			\hline
			WHDR & 25.70\% &  31.31\% & {\bf 14.39\%} & 22.14\% & 31.37\%\\
			\hline
		\end{tabular}
	\end{center}
	\caption{Weighted Human Disagreement Rate (WHDR) of various methods on our DIW
          dataset, including Eigen(V), the method of Eigen et
          al.~\cite{eigen2015predicting} (VGGNet~\cite{simonyan2014very} version) }
	\label{table:amt_performance}
\end{table}

\section{Conclusions}
We have studied single-image depth perception in the wild, recovering depth from a single image taken in
unconstrained settings. We have introduced a new dataset consisting of images in the wild annotated with
relative depth and proposed a new algorithm that learns to
estimate metric depth supervised by relative depth. We have shown that our
algorithm outperforms prior art and our algorithm,
combined with existing RGB-D data and our new relative depth annotations, significantly
improves single-image depth perception in the wild. 

\vspace{-2 mm}
\subsubsection*{Acknowledgments}
\vspace{-2 mm}
This work is partially supported by the National Science Foundation under Grant No. 1617767.

\vspace{-2 mm}
\bibliographystyle{ieeetr}
{\small\bibliography{arxiv_v2}}

\newpage 

\begin{center}
\textbf{{\LARGE Appendix}}
\end{center}

\begin{figure}[h]
	\centering
	\includegraphics[width=1.0\textwidth]{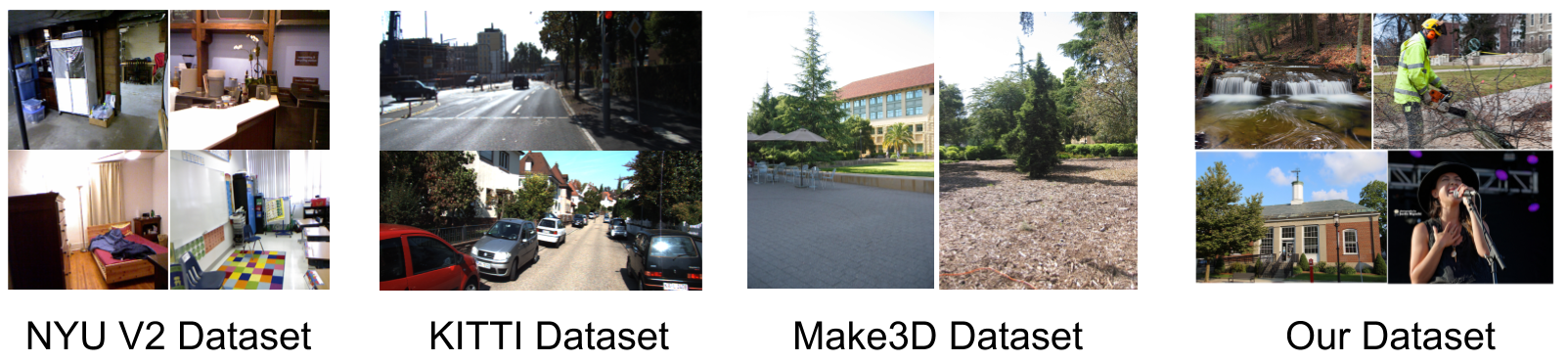}
	\caption{Additional example images from current RGB-D datasets and our Depth in the Wild
		(DIW) dataset.}
	\label{fig:dataset_comparison_suppl}
\end{figure}

\begin{figure}[h] 
	\centering
	\includegraphics[width=1.0\textwidth]{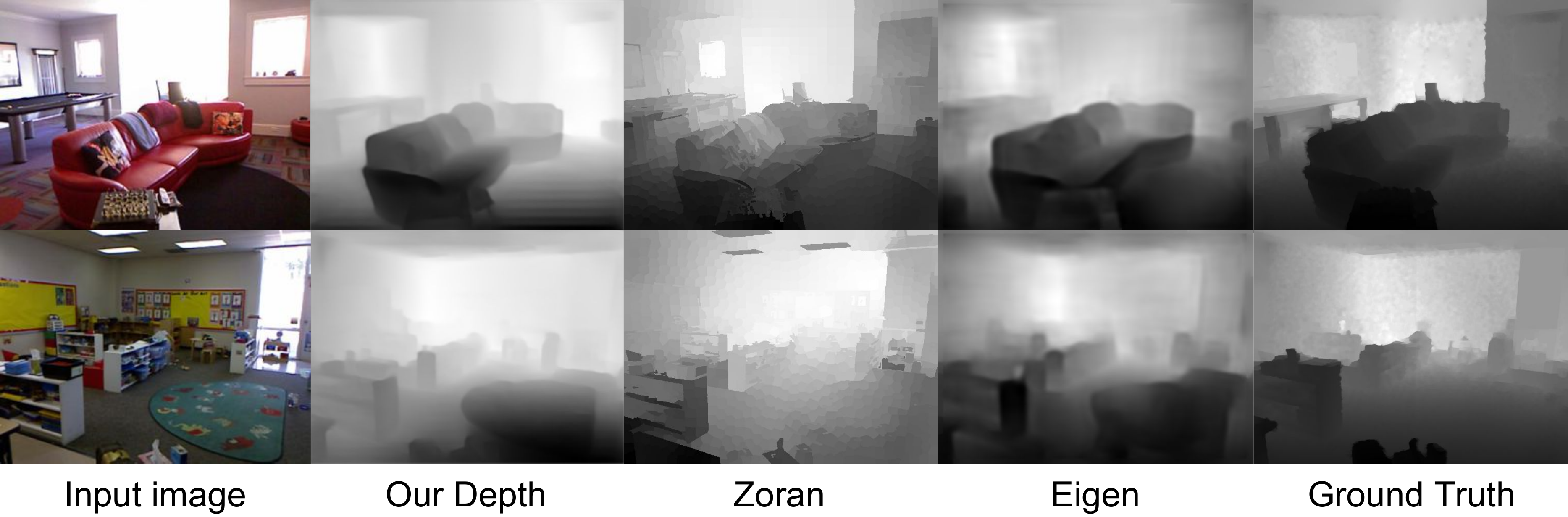}
	\caption{Additional qualitative results on NYU Depth by our method, the method of Eigen et
		al.~\cite{eigen2015predicting}, and the method of Zoran et
		al.~\cite{zoran2015learning}. All depth maps except ours are directly from \cite{zoran2015learning}. }
	\label{fig:qual_nyu_suppl}
\end{figure}

\begin{figure}[t] 
	\centering
	\includegraphics[width=1.0\textwidth]{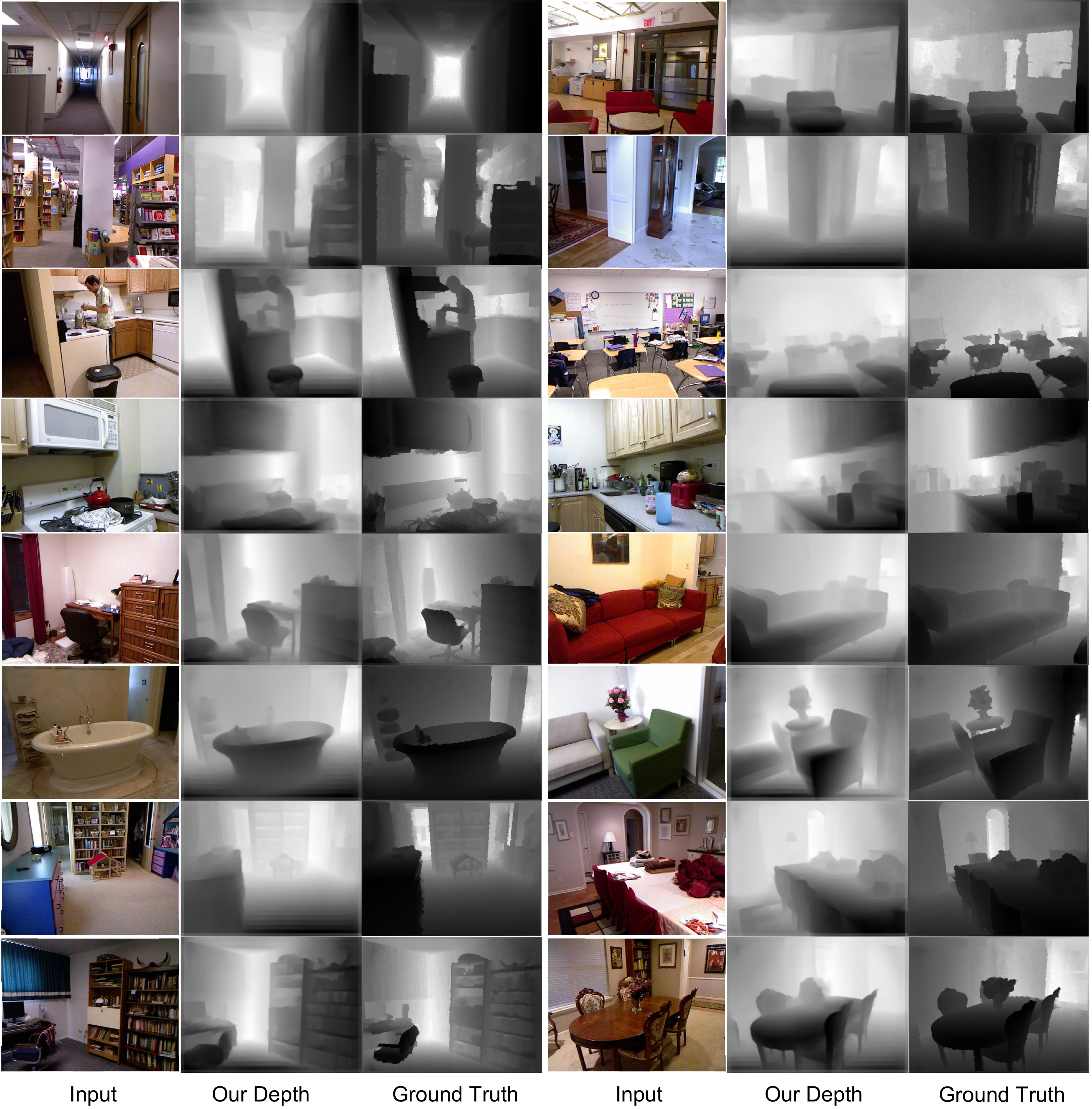}
	\caption{Additional qualitative results on NYU Depth test set by our method. Here we show the original input images and the depth maps by our method, as well as the ground truth.}
	\label{fig:more_qual_nyu}
\end{figure}

\begin{figure}[t]
	\centering
	\includegraphics[width=0.8\textwidth]{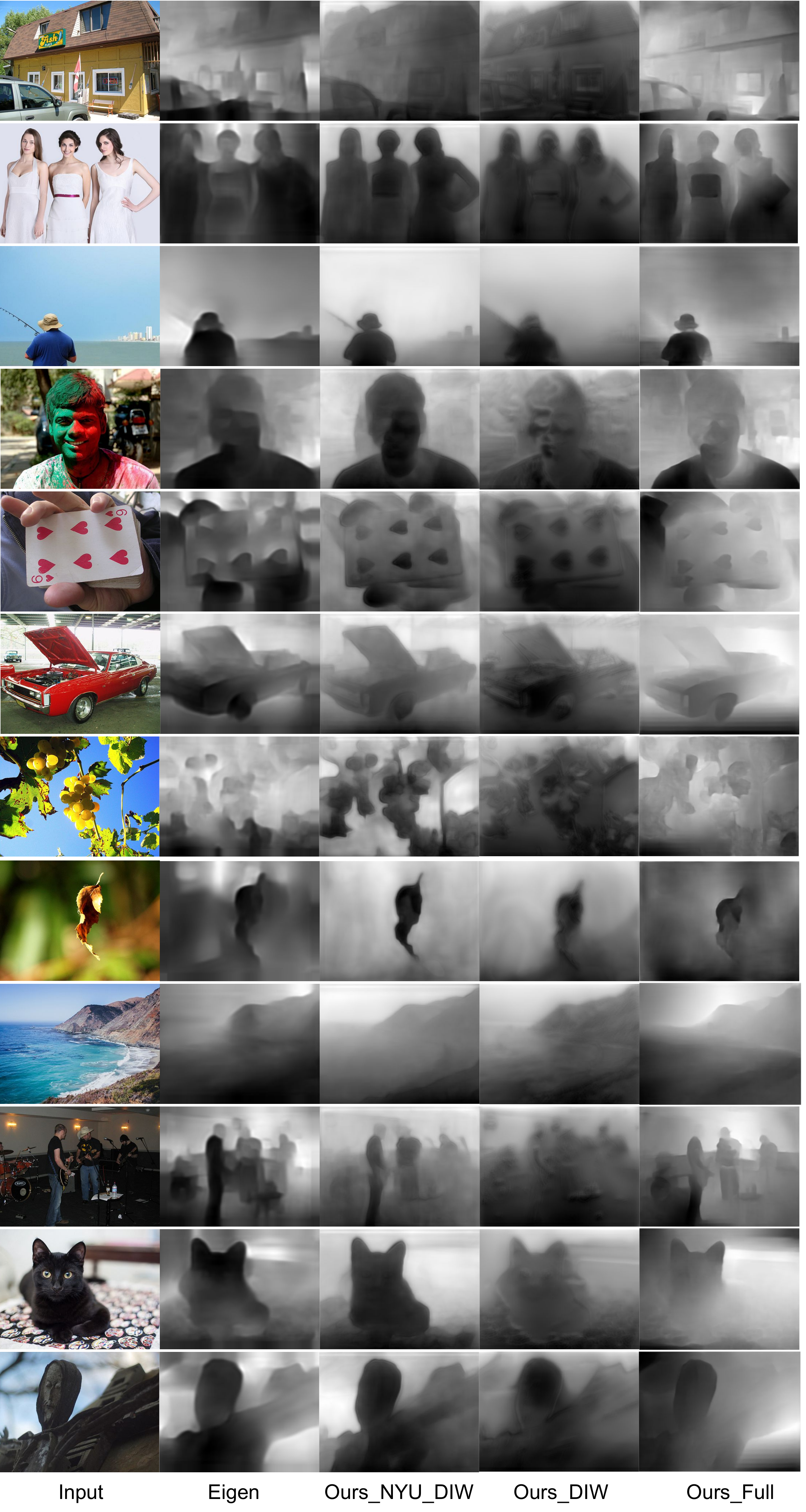}
	\caption{Additional qualitative results on our Depth in the Wild (DIW) dataset by our method and the method of Eigen et al.~\cite{eigen2015predicting}.}
	\label{fig:qual_DIW_suppl}
\end{figure}

\clearpage

\begin{table}[t]
	\centering
	\begin{center}
		\begin{tabular}{  l  c  c  c  c  c } 
			\toprule
			Method & RMSE & RMSE & RMSE   & absrel & sqrrel \\
			&      & (log)&(s.inv) &        &        \\
			\midrule
			Ours & 0.89 & 0.32 & 0.25   & 0.27  &  0.29  \\
			Ours\_Full & 0.74 & 0.26 & 0.21   & 0.21  &  0.19  \\
			Eigen(V)~\cite{eigen2015predicting} & \textbf{0.64} & \textbf{0.21} & \textbf{0.17} & \textbf{0.16} & \textbf{0.12} \\
			\bottomrule
		\end{tabular}
	\end{center}
	\caption{ Table 2 of the main paper reports the metric error of our network trained on relative depth pairs. Here we provide additional results by training our network on the full depth map. The network \textbf{Ours} is our network trained on the 795 NYU Depth training subset, and \textbf{Ours\_Full} is our network trained on the full NYU Depth training set.} 
	\label{table:full_depthmap_result}
\end{table}

\begin{table}[t]
	\centering
	\begin{center}
		\begin{tabular}{  l  c  c  c } 
			\toprule
			Method & WKDR & WKDR$^{=}$  & WKDR$^{\neq}$ \\
			\midrule
			rand\_48K  &  {\bf 34.3\% } & {\bf 31.7\%} & {\bf 37.1\%}\\
			rand\_24K  &  34.5\% & 32.6\% & 36.9\%\\
			rand\_12K  &  34.9\% & 32.4\% & 37.6\%\\
			rand\_6K  &   36.1\% & 32.2\% & 39.9\%\\
			rand\_3K  &   35.8\% & 28.7\% & 41.3\%\\
			\bottomrule
		\end{tabular}
	\end{center}
	\caption{ Table 2 of the main paper reports the performance of our network versus the number of randomly sampled non-superpixel point pairs on NYU Depth. Here we report additional results by sampling more pairs. $\mathbf{rand\_N}$ denotes a network trained with $N$ pairs per image.}
	\label{table:random_points}
\end{table}

\begin{table}[h]
	\centering
	\begin{center}
		\begin{tabular}{  c  c  c  c } 
			\toprule
			\#Depth Pairs & WKDR & WKDR$^{=}$  & WKDR$^{\neq}$ \\
			\midrule
			800  &  {\bf 35.6\% }& {\bf 36.1\% } & {\bf 36.5\%} \\
			500  &   37.2\% & 37.7\% & 38.2\%\\
			250  &   38.0\% & 37.4\% & 39.7\%\\
			100  &   42.3\% & 41.1\% & 44.0\%\\
			\bottomrule
		\end{tabular}
	\end{center}
	\caption{ Table 2 of the main paper reports the performance of our network trained on 800 superpixel point pairs. Here we report additional results by decreasing the number of point pairs. }
	\label{table:structured_points}
\end{table}

\end{document}